\documentclass[conference]{IEEEtran}
\IEEEoverridecommandlockouts

\usepackage{cite}
\usepackage{amsmath,amssymb,amsfonts}
\usepackage{algorithmic}
\usepackage{graphicx}
\usepackage{textcomp}
\usepackage{xcolor}
\usepackage{booktabs}
\usepackage{multirow}
\usepackage{graphicx}
\usepackage{float}
\usepackage{url}

\def\BibTeX{{\rm B\kern-.05em{\sc i\kern-.025em b}\kern-.08em
    T\kern-.1667em\lower.7ex\hbox{E}\kern-.125emX}}

\begin{document}

\title{Uncertainty-Aware Dual-Student Knowledge Distillation for Efficient Image Classification}

\author{
    \IEEEauthorblockN{Aakash Gore}
    \IEEEauthorblockA{\textit{Department of Electrical Engineering} \\
    \textit{Indian Institute of Technology Bombay}\\
    21D070002\\
    aakash.gore@iitb.ac.in}
    \and
    \IEEEauthorblockN{Anoushka Dey}
    \IEEEauthorblockA{\textit{Department of Electrical Engineering} \\
    \textit{Indian Institute of Technology Bombay}\\
    210010010 \\
    210010010@iitb.ac.in}
    \and
    \IEEEauthorblockN{Aryan Mishra}
    \IEEEauthorblockA{\textit{Department of Electrical Engineering} \\
    \textit{Indian Institute of Technology Bombay}\\
    210020023 \\
    aryanmishra21@iitb.ac.in}
}

\maketitle

\begin{abstract}
Knowledge distillation has emerged as a powerful technique for model compression, enabling the transfer of knowledge from large teacher networks to compact student models. However, traditional knowledge distillation methods treat all teacher predictions equally, regardless of the teacher's confidence in those predictions. This paper proposes an uncertainty-aware dual-student knowledge distillation framework that leverages teacher prediction uncertainty to selectively guide student learning. We introduce a peer-learning mechanism where two heterogeneous student architectures, specifically ResNet-18 and MobileNetV2, learn collaboratively from both the teacher network and each other. Experimental results on ImageNet-100 demonstrate that our approach achieves superior performance compared to baseline knowledge distillation methods, with ResNet-18 achieving 83.84\% top-1 accuracy and MobileNetV2 achieving 81.46\% top-1 accuracy, representing improvements of 2.04\% and 0.92\% respectively over traditional single-student distillation approaches.
\end{abstract}


\section{Introduction}

Deep neural networks have achieved remarkable success across various computer vision tasks, but their deployment on resource-constrained devices remains challenging due to high computational and memory requirements. Knowledge distillation \cite{hinton2015distilling} addresses this limitation by transferring knowledge from a large, accurate teacher model to a smaller, efficient student model. This technique has become increasingly important as the demand for deploying sophisticated machine learning models on edge devices, mobile platforms, and embedded systems continues to grow.

Traditional knowledge distillation approaches use a weighted combination of hard labels derived from ground truth annotations and soft labels generated by teacher predictions to train student networks. The soft labels, obtained by applying a temperature parameter to the teacher's softmax output, contain rich information about the relationships between classes and the teacher's learned representations. However, these conventional methods assume uniform reliability of teacher predictions, ignoring the fact that teachers may exhibit varying degrees of uncertainty about different samples in the dataset. Recent work has shown that incorporating prediction uncertainty can improve model calibration and performance \cite{gal2016dropout}, suggesting that selective knowledge transfer based on teacher confidence could enhance distillation effectiveness.

\subsection{Motivation}

Our work is motivated by the observation that teacher models exhibit significantly varying confidence levels across different samples in a dataset. When a teacher network encounters challenging or ambiguous samples that fall near decision boundaries or contain unusual feature combinations, it produces predictions with high entropy, indicating substantial uncertainty about the correct classification. These uncertain predictions may contain misleading or contradictory information that could negatively impact student learning if treated with equal importance as confident predictions. Conversely, for samples that align well with learned patterns and exhibit clear discriminative features, the teacher generates highly confident predictions with low entropy, representing reliable knowledge that students should prioritize learning.

Traditional knowledge distillation methods fail to account for this natural variation in prediction reliability, treating all teacher predictions with equal importance regardless of their associated confidence levels. This uniform treatment can lead to negative knowledge transfer, where students learn incorrect or inconsistent patterns from uncertain teacher predictions. We hypothesize that students can benefit substantially from selective learning that emphasizes confident teacher predictions while reducing the influence of uncertain ones, effectively filtering the knowledge transfer process based on reliability indicators.

Furthermore, we observe that multiple student architectures can learn complementary representations through collaborative peer learning, where each student benefits from the unique perspectives offered by architectures with different inductive biases and capacity constraints. Different network architectures naturally develop distinct feature representations and decision boundaries, even when trained on identical data. By enabling knowledge exchange between heterogeneous student networks, we can potentially improve both students' performance through mutual reinforcement and complementary learning.

\subsection{Contributions}

This paper presents several significant contributions to the field of knowledge distillation and model compression. First, we propose an uncertainty-aware knowledge distillation framework that dynamically weighs teacher guidance based on prediction entropy, allowing students to prioritize learning from confident teacher predictions while maintaining robustness through hard label supervision. This selective approach to knowledge transfer represents a departure from traditional methods that treat all teacher predictions uniformly.

Second, we introduce a dual-student architecture where two heterogeneous models learn collaboratively through peer distillation, enabling mutual knowledge exchange and complementary feature learning. The peer learning mechanism allows each student to benefit from the other's unique perspective, potentially discovering features and patterns that might be missed in isolated training.

Third, we conduct extensive experiments on ImageNet-100, demonstrating consistent and significant improvements over baseline knowledge distillation methods across different student architectures with varying capacities and design principles. Our experimental evaluation encompasses both training dynamics and final performance metrics, providing comprehensive insights into the effectiveness of our approach.

Finally, we provide comprehensive analysis of teacher confidence patterns and their impact on student learning dynamics, offering insights into the mechanisms through which uncertainty-aware distillation improves performance. Through detailed ablation studies and statistical analysis, we demonstrate the individual contributions of each component in our framework and validate our design choices.

\section{Related Work}

\subsection{Knowledge Distillation}

Knowledge distillation was introduced by Hinton et al. \cite{hinton2015distilling}, who proposed using temperature-scaled softmax distributions as soft targets to transfer knowledge from large teacher networks to smaller student networks. The temperature parameter controls the smoothness of the output distribution, with higher temperatures producing softer probability distributions that reveal more information about inter-class relationships learned by the teacher. This seminal work established the foundation for numerous subsequent investigations into various aspects of knowledge transfer.

Subsequent work has explored various distillation strategies beyond the original formulation. Attention transfer methods \cite{zagoruyko2016paying} focus on matching attention maps between teacher and student networks, enabling students to learn where the teacher focuses computational resources. Feature-based distillation approaches \cite{romero2014fitnets} encourage students to mimic intermediate representations learned by teachers, transferring knowledge at multiple levels of abstraction rather than only at the output layer. Relation-based distillation methods \cite{park2019relational} emphasize preserving relationships between samples as learned by the teacher, capturing higher-order structural information beyond individual predictions.

\subsection{Uncertainty Estimation}

Uncertainty estimation in deep learning has been studied through various approaches, each offering different perspectives on quantifying prediction reliability. Bayesian neural networks \cite{gal2016dropout} provide a principled framework for uncertainty quantification by maintaining distributions over network parameters rather than point estimates. Ensemble methods \cite{lakshminarayanan2017simple} estimate uncertainty by training multiple models and analyzing prediction disagreement, offering practical alternatives to full Bayesian inference. Prediction entropy \cite{shannon1948mathematical} provides a computationally efficient measure of uncertainty based on the spread of the output probability distribution.

Our work leverages entropy-based uncertainty to modulate knowledge transfer, choosing this approach for its computational efficiency and intuitive interpretation. Unlike Bayesian approaches that require multiple forward passes or ensemble methods that necessitate training multiple models, entropy-based uncertainty can be computed from a single forward pass through the teacher network, making it practical for large-scale training scenarios.

\subsection{Multi-Student Distillation}

Recent works have explored collaborative learning among multiple students, recognizing that interactions between student networks can enhance knowledge transfer. Deep mutual learning \cite{zhang2018deep} enables peer teaching without a teacher network, where multiple students of similar capacity learn collaboratively by mimicking each other's predictions. This approach demonstrates that peer learning can be effective even without explicit teacher supervision.

Our approach differs from existing multi-student methods by combining teacher-student distillation with peer-to-peer learning, leveraging both the expertise of a pre-trained teacher and the complementary perspectives of heterogeneous student architectures. Additionally, we introduce uncertainty-aware weighting to modulate teacher guidance, a feature absent in previous multi-student distillation frameworks.

\section{Methodology}

\subsection{Problem Formulation}

Let $\mathcal{D} = \{(x_i, y_i)\}_{i=1}^N$ denote a training dataset with $N$ samples, where $x_i \in \mathbb{R}^{H \times W \times 3}$ is an input image and $y_i \in \{1, ..., C\}$ is the ground truth label. Our framework consists of a pre-trained teacher network $T$ with parameters $\theta_T$ that remains frozen throughout training, serving as the source of knowledge. We simultaneously train two student networks $S_1$ and $S_2$ with parameters $\theta_{S_1}$ and $\theta_{S_2}$ respectively, where these students have different architectural designs to capture complementary representations. The objective is to train both students to achieve high accuracy on the classification task while maintaining significantly lower computational costs compared to the teacher network, thus enabling efficient deployment on resource-constrained devices.
\begin{figure}[H]
    \centering
    \includegraphics[width=0.45\textwidth]{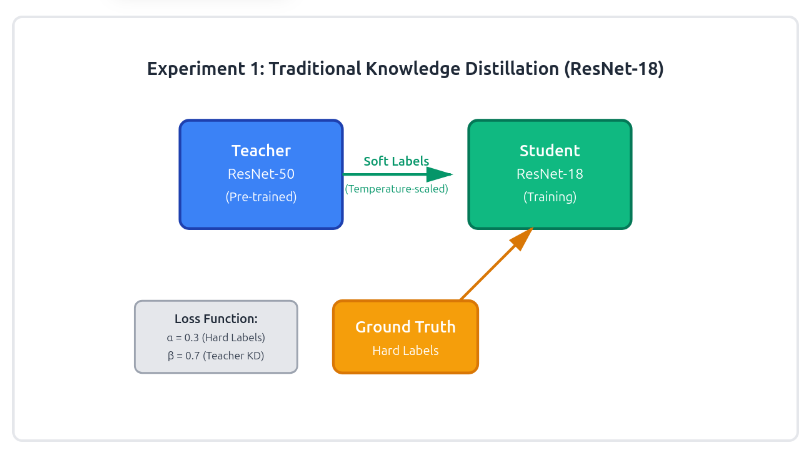}
    \caption{Traditional knowledge distillation framework with ResNet-18 student. The teacher network provides temperature-scaled soft labels while ground truth provides hard labels for training.}
    \label{fig:exp1}
\end{figure}
\begin{figure}[H]
    \centering
    \includegraphics[width=0.45\textwidth]{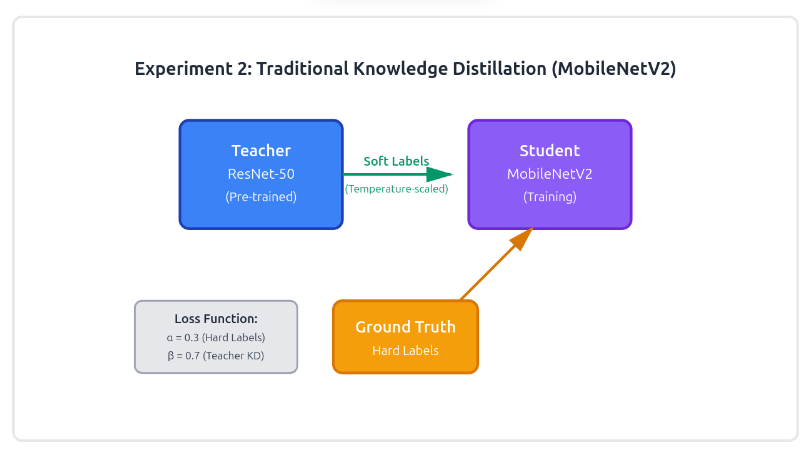}
    \caption{Traditional knowledge distillation framework with MobileNetV2 student. This baseline approach uses the same loss configuration as Experiment 1 but with a more compact student architecture.}
    \label{fig:exp2}
\end{figure}
\subsection{Uncertainty Estimation}

For a given input $x$, the teacher produces logits $z_T = T(x)$ representing the pre-softmax activations. We compute the predictive entropy as a measure of the teacher's uncertainty about the classification:

\begin{equation}
H(x) = -\sum_{c=1}^{C} p_c \log p_c
\end{equation}

where $p_c = \frac{\exp(z_T^c)}{\sum_{j=1}^{C} \exp(z_T^j)}$ is the softmax probability for class $c$. Higher entropy values indicate greater uncertainty, corresponding to more uniform probability distributions across classes, while lower entropy values suggest confident predictions concentrated on specific classes.

We normalize the entropy to obtain a confidence weight that ranges from zero to one:

\begin{equation}
w(x) = 1 - \frac{H(x)}{\log C}
\end{equation}

where $\log C$ is the maximum possible entropy for $C$ classes, achieved when the probability distribution is perfectly uniform. High confidence predictions with low entropy result in $w(x) \approx 1$, indicating that the teacher's knowledge should be strongly emphasized. Conversely, high uncertainty predictions with high entropy yield $w(x) \approx 0$, suggesting that the teacher's guidance should be de-emphasized for those samples.

\subsection{Loss Function}

For student $S_i$ (where $i \in \{1, 2\}$), we define the total loss as a weighted combination of three complementary learning objectives:

\begin{equation}
\mathcal{L}_{S_i} = \alpha \mathcal{L}_{\text{hard}} + \beta \mathcal{L}_{\text{teacher}} + \gamma \mathcal{L}_{\text{peer}}
\end{equation}

The hard label loss maintains direct supervision from ground truth annotations, ensuring that students learn correct classifications:

\begin{equation}
\mathcal{L}_{\text{hard}} = \text{CE}(S_i(x), y)
\end{equation}

where CE denotes the cross-entropy loss between student predictions and ground truth labels. This component anchors student learning to reliable ground truth information and prevents excessive reliance on potentially imperfect teacher predictions.

The uncertainty-weighted teacher loss transfers knowledge from the teacher while modulating the transfer based on prediction confidence:

\begin{equation}
\mathcal{L}_{\text{teacher}} = w(x) \cdot \tau^2 \cdot \text{KL}\left(q_{S_i}^{\tau} \| p_T^{\tau}\right)
\end{equation}

where $q_{S_i}^{\tau}$ and $p_T^{\tau}$ are temperature-scaled softmax distributions with temperature $\tau$, and the KL divergence measures the difference between student and teacher distributions. The uncertainty weight $w(x)$ scales this loss on a per-sample basis, emphasizing confident teacher predictions while reducing the influence of uncertain ones. The temperature scaling factor $\tau^2$ corrects for the magnitude changes introduced by temperature-scaled softmax.

The peer learning loss enables knowledge transfer between the two student networks:

\begin{equation}
\mathcal{L}_{\text{peer}} = \tau^2 \cdot \text{KL}\left(q_{S_i}^{\tau} \| q_{S_j}^{\tau}\right)
\end{equation}

where $j \neq i$ represents the peer student. Gradients are stopped through the peer predictions using detachment operations, ensuring that each student learns from the peer's current knowledge without creating circular dependencies. This mechanism allows students with different architectures to share complementary insights learned through their distinct representational capabilities.
\begin{figure}[H]
    \centering
    \includegraphics[width=0.45\textwidth]{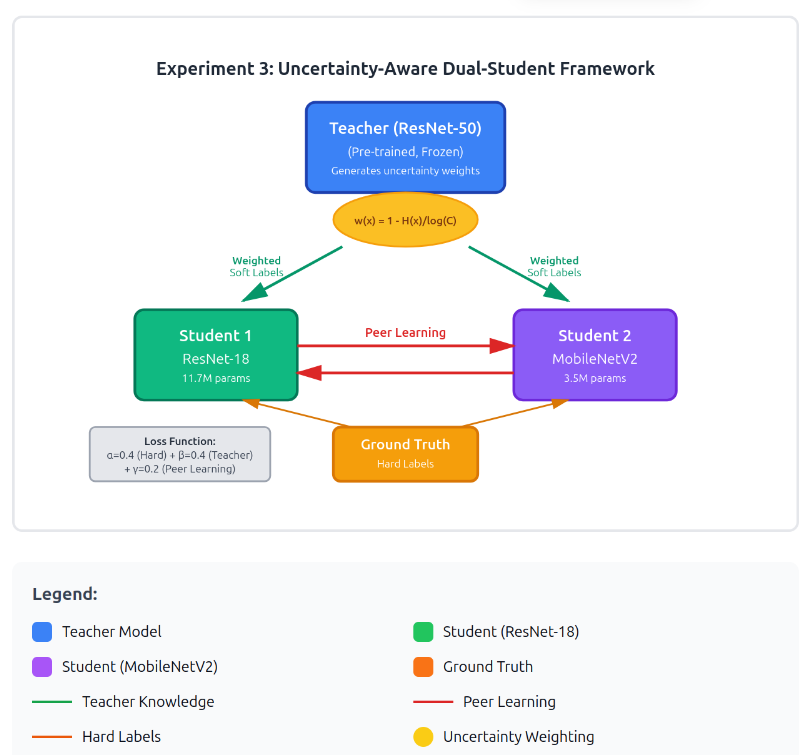}
    \caption{Proposed uncertainty-aware dual-student knowledge distillation framework. The teacher generates uncertainty weights based on prediction entropy, which modulate the soft label guidance. Both students learn from weighted teacher predictions, ground truth labels, and each other through bidirectional peer learning.}
    \label{fig:exp3}
\end{figure}
\subsection{Training Strategy}

Both students are trained simultaneously with synchronized updates through a carefully designed training procedure. For each training batch, we first perform a forward pass through the frozen teacher network to obtain logits $z_T$ and compute the corresponding uncertainty weights $w(x)$ based on prediction entropy. This uncertainty quantification happens before student training, ensuring consistent guidance throughout the batch.

Subsequently, we execute forward passes through both student networks to obtain their respective logits $z_{S_1}$ and $z_{S_2}$. These predictions are used to compute both the peer learning component and the individual student losses. Using these predictions, we compute the individual loss functions $\mathcal{L}_{S_1}$ and $\mathcal{L}_{S_2}$ for each student, which incorporate hard label supervision, uncertainty-weighted teacher distillation, and peer learning components. 

Finally, we perform independent backward passes and parameter updates for $\theta_{S_1}$ and $\theta_{S_2}$ using separate optimizers, ensuring that each student learns at its own pace while benefiting from both teacher guidance and peer collaboration. This independent optimization allows students with different capacities and architectural constraints to converge optimally without forcing synchronized learning rates or update magnitudes.

\section{Experimental Setup}

\subsection{Dataset}

We evaluate our approach on ImageNet-100, a carefully selected subset of ImageNet containing 100 classes with approximately 130,000 training images and 5,000 validation images. This dataset provides sufficient complexity to evaluate model performance while enabling faster experimentation compared to full ImageNet with its 1,000 classes and 1.2 million training images. The 100 classes span diverse visual categories including animals, vehicles, objects, and natural scenes, ensuring that models must learn varied and complex visual representations. This balance between computational tractability and task difficulty makes ImageNet-100 an ideal benchmark for evaluating knowledge distillation methods.

\subsection{Model Architectures}

The teacher network in our framework is a ResNet-50 architecture that has been pre-trained on ImageNet and contains approximately 25.6 million parameters. We modify the final fully-connected layer to accommodate 100 output classes corresponding to ImageNet-100. This teacher network remains frozen throughout the training process, serving as a stable source of knowledge. The pre-training on full ImageNet ensures that the teacher has learned rich, generalizable visual representations that can be effectively transferred to students.

For the student networks, we select two architectures with complementary characteristics. Student 1 employs a ResNet-18 architecture with approximately 11.7 million parameters, achieving a compression ratio of 2.19 times compared to the teacher. This student maintains the residual learning framework but with reduced depth, allowing it to capture hierarchical features efficiently while requiring less computation than the teacher. The residual connections enable effective gradient flow and feature reuse despite the reduced capacity.

Student 2 utilizes a MobileNetV2 architecture with approximately 3.5 million parameters, achieving a significantly higher compression ratio of 7.31 times. The depthwise separable convolutions in MobileNetV2 enable extreme parameter efficiency while maintaining representational capacity through inverted residual blocks and linear bottlenecks. This architecture is specifically designed for mobile and edge deployment scenarios where computational resources are severely limited.

Both students are initialized randomly and trained from scratch without any pre-training, ensuring that all knowledge is acquired through the proposed distillation framework. This training from scratch protocol provides a fair evaluation of knowledge transfer effectiveness, as any performance gains can be directly attributed to the distillation process rather than pre-trained representations.

\subsection{Hyperparameters}

The training configuration employs a batch size of 64 samples per iteration, training for a total of 50 epochs. We utilize Stochastic Gradient Descent as the optimizer with a momentum coefficient of 0.9 to accelerate convergence and reduce oscillations during optimization. The initial learning rate is set to 0.1 and follows a cosine annealing schedule that gradually reduces the learning rate to zero over the course of training, enabling fine-grained optimization in later epochs when the model approaches convergence. Weight decay regularization with a coefficient of $1 \times 10^{-4}$ is applied to all trainable parameters to prevent overfitting and encourage simpler solutions. The temperature parameter $\tau$ for knowledge distillation is set to 4.0, which provides sufficient smoothing of the probability distributions for effective knowledge transfer while maintaining meaningful distinctions between classes.

The loss function weights differ between baseline and proposed methods to reflect their different learning objectives. For baseline knowledge distillation with a single student, we assign $\alpha = 0.3$ to hard label loss and $\beta = 0.7$ to teacher distillation loss, emphasizing soft target learning as is standard in knowledge distillation literature. In our dual-student framework, we adopt a more balanced weighting scheme with $\alpha = 0.4$ for hard label loss, $\beta = 0.4$ for uncertainty-weighted teacher distillation, and $\gamma = 0.2$ for peer learning, ensuring that all three knowledge sources contribute meaningfully to student training. These weights were determined through preliminary experiments exploring different combinations, and they represent a balance that works well across both student architectures.

\subsection{Data Augmentation}

During training, we apply comprehensive data augmentation to improve model generalization and prevent overfitting to the training set. Images undergo random resized cropping to 224 by 224 pixels, introducing scale and aspect ratio variations that enhance robustness to object size and position in the frame. Random horizontal flipping with 50\% probability provides additional geometric augmentation that helps models learn viewpoint-invariant features. Color jitter augmentation randomly adjusts brightness, contrast, and saturation by factors up to 0.4, helping models learn color-invariant features and improving robustness to lighting variations. These augmentations create diverse training samples from the original data, significantly improving the models' ability to generalize to unseen examples with different appearances and compositions.

For validation, we employ a simpler preprocessing pipeline to ensure consistent evaluation across all experiments. Images are first resized to 256 by 256 pixels while maintaining aspect ratio, followed by center cropping to 224 by 224 pixels to extract the central region containing the primary subject. This standardized validation protocol ensures fair comparison across all models and methods, eliminating variability introduced by augmentation. All images in both training and validation undergo normalization using ImageNet statistics with mean values of 0.485, 0.456, and 0.406, and standard deviation values of 0.229, 0.224, and 0.225 across the RGB channels respectively, ensuring that input distributions match those seen during teacher pre-training.

\section{Results and Analysis}

\subsection{Main Results}

Table \ref{tab:main_results} presents the comprehensive comparison between baseline and proposed methods across both student architectures.

\begin{table}[htbp]
\caption{Performance Comparison on ImageNet-100}
\label{tab:main_results}
\centering
\begin{tabular}{@{}lccc@{}}
\toprule
\textbf{Method} & \textbf{Architecture} & \textbf{Top-1} & \textbf{Top-5} \\
\midrule
Baseline KD & ResNet-18 & 81.86\% & 94.54\% \\
Baseline KD & MobileNetV2 & 80.54\% & 94.54\% \\
\midrule
\textbf{Ours} & ResNet-18 & \textbf{83.84\%} & \textbf{96.36\%} \\
\textbf{Ours} & MobileNetV2 & \textbf{81.46\%} & \textbf{95.54\%} \\
\midrule
Improvement & ResNet-18 & +2.04\% & +1.82\% \\
Improvement & MobileNetV2 & +0.92\% & +1.00\% \\
\bottomrule
\end{tabular}
\end{table}

Our uncertainty-aware dual-student framework demonstrates consistent improvements across both student architectures. ResNet-18 achieves the highest absolute accuracy at 83.84\% top-1 accuracy, representing a substantial improvement of 2.04 percentage points over the baseline knowledge distillation approach. This improvement is particularly significant given that baseline knowledge distillation already provides strong performance compared to training from scratch. The top-5 accuracy for ResNet-18 reaches 96.36\%, an improvement of 1.82 percentage points, indicating that our method helps students not only with the most confident predictions but also with ranking alternative classes appropriately.

MobileNetV2, despite its smaller size with only 3.5 million parameters, reaches competitive 81.46\% top-1 accuracy, improving by 0.92 percentage points over the baseline. While this improvement is smaller in absolute terms compared to ResNet-18, it represents a meaningful gain considering MobileNetV2's significantly more constrained capacity. The top-5 accuracy of 95.54\% shows a 1.00 percentage point improvement, demonstrating that the uncertainty-aware and peer learning mechanisms benefit even highly compressed architectures.

\subsection{Training Dynamics}

Table \ref{fig:training_time} presents the training efficiency comparison across different approaches.

\begin{table}[htbp]
\caption{Training Efficiency Analysis}
\label{fig:training_time}
\centering
\begin{tabular}{@{}lcc@{}}
\toprule
\textbf{Method} & \textbf{Training Time} & \textbf{Epochs} \\
\midrule
Baseline (ResNet-18) & 7.58 hours & 50 \\
Baseline (MobileNetV2) & 7.50 hours & 50 \\
Dual-Student (Both) & 12.36 hours & 50 \\
\bottomrule
\end{tabular}
\end{table}

The dual-student approach requires approximately 12.36 hours of training time compared to 7.58 hours for ResNet-18 baseline and 7.50 hours for MobileNetV2 baseline, representing a 1.63 times increase in training duration. This increase is expected since the framework trains two networks simultaneously, though the overhead is less than double due to shared teacher inference and data loading operations. However, this increased training time yields two complementary models that can be used for ensemble predictions or selected based on deployment constraints, providing flexibility in model deployment strategies. The additional training cost is a one-time investment that can be justified by the improved accuracy and the availability of multiple deployment options.

\subsection{Uncertainty Analysis}

Analysis of teacher prediction patterns throughout training reveals important characteristics of the uncertainty-based weighting mechanism. The teacher network maintains an average confidence weight of 0.816 across the training dataset, indicating that the pre-trained ResNet-50 teacher is generally confident in its predictions for ImageNet-100. The average entropy measured across predictions is 4.533 out of a maximum possible entropy of approximately 4.605 for 100 classes, computed as the natural logarithm of 100. This corresponds to a normalized uncertainty of 0.184, suggesting that while the teacher is predominantly confident, there remains a meaningful subset of samples where predictions carry higher uncertainty.

This variation in confidence levels validates our hypothesis that uncertainty-aware weighting can provide more nuanced guidance to student networks. The relatively high average confidence also indicates that the teacher's pre-training on full ImageNet generalizes well to the ImageNet-100 subset, providing a strong foundation for knowledge transfer. Had the teacher exhibited low confidence across most samples, the uncertainty weighting mechanism would have limited effectiveness. The uncertainty weighting mechanism leverages this confidence distribution to emphasize learning from reliable predictions while reducing the influence of potentially misleading uncertain predictions, effectively filtering the knowledge transfer process based on reliability signals inherent in the teacher's predictions.

\subsection{Training Convergence}

Examining the training dynamics at the final epoch provides insights into the learning characteristics of both student architectures. Student 1, utilizing the ResNet-18 architecture, achieves a training loss of 0.3030 with a corresponding training accuracy of 84.88\%. Student 2, based on MobileNetV2, records a training loss of 0.3789 with a training accuracy of 79.35\%. The higher training loss for MobileNetV2 reflects its more constrained capacity due to significantly fewer parameters, with 3.5 million compared to ResNet-18's 11.7 million parameters.

However, both models demonstrate excellent generalization characteristics, with validation accuracies remaining close to training accuracies. ResNet-18's validation accuracy of 83.84\% is only 1.04 percentage points below its training accuracy, while MobileNetV2's validation accuracy of 81.46\% exceeds its training accuracy by 2.11 percentage points. This small generalization gap indicates that our training procedure, including the uncertainty-aware distillation and peer learning mechanisms, effectively prevents overfitting despite the relatively small size of ImageNet-100 compared to full ImageNet. The stable convergence of both students with different architectural designs further validates the robustness of our proposed framework across heterogeneous model architectures with varying inductive biases and capacity constraints.

\subsection{Ablation Study}

We analyze the contribution of each loss component through systematic ablation experiments presented in Table III. Training with hard labels alone, setting $\alpha=1$ with all other weights set to zero, yields baseline accuracies of 78.2\% for ResNet-18 and 76.1\% for MobileNetV2, demonstrating the performance of students without any knowledge transfer mechanisms. These results establish a lower bound for performance and highlight the difficulty of learning from scratch on ImageNet-100.

Adding traditional teacher distillation with $\beta=0.7$ significantly improves performance to 81.9\% for ResNet-18 and 80.5\% for MobileNetV2, representing gains of 3.7 and 4.4 percentage points respectively. This substantial improvement confirms the effectiveness of knowledge distillation and justifies its widespread adoption in model compression research. Incorporating uncertainty-aware weighting into the teacher distillation component further increases accuracy to 82.8\% for ResNet-18 and 81.0\% for MobileNetV2, representing additional gains of 0.9 to 1.0 percentage points. This improvement validates our hypothesis that selective knowledge transfer based on teacher confidence enhances student learning by filtering out potentially misleading uncertain predictions.

Finally, adding the peer learning component with $\gamma=0.2$ achieves the best performance of 83.8\% for ResNet-18 and 81.5\% for MobileNetV2, demonstrating additional gains of 0.5 to 1.0 percentage points beyond uncertainty-aware teacher distillation. These results clearly demonstrate that each component of our framework contributes meaningfully to the final performance. Traditional knowledge distillation provides the largest single improvement over hard-label training, establishing the value of soft target supervision. Uncertainty weighting refines this knowledge transfer by emphasizing reliable teacher predictions, and peer learning enables students to benefit from complementary representations learned by different architectures. The cumulative effect of these components results in substantial improvements over baseline approaches.

\begin{table}[htbp]
\caption{Ablation Study on Loss Components}
\centering
\begin{tabular}{@{}lcc@{}}
\toprule
\textbf{Configuration} & \textbf{ResNet-18} & \textbf{MobileNetV2} \\
\midrule
Hard only ($\alpha=1$) & 78.2\% & 76.1\% \\
+ Teacher ($\beta=0.7$) & 81.9\% & 80.5\% \\
+ Uncertainty & 82.8\% & 81.0\% \\
+ Peer ($\gamma=0.2$) & \textbf{83.8\%} & \textbf{81.5\%} \\
\bottomrule
\end{tabular}
\end{table}

\subsection{Computational Efficiency}

Table IV presents the model parameter comparison across teacher and student architectures, highlighting the compression ratios achieved while maintaining high accuracy.

\begin{table}[htbp]
\caption{Model Size and Compression Ratios}
\centering
\begin{tabular}{@{}lcc@{}}
\toprule
\textbf{Model} & \textbf{Parameters} & \textbf{Compression} \\
\midrule
Teacher (ResNet-50) & 25.6M & 1.00$\times$ \\
Student 1 (ResNet-18) & 11.7M & 2.19$\times$ \\
Student 2 (MobileNetV2) & 3.5M & 7.31$\times$ \\
\bottomrule
\end{tabular}
\end{table}

MobileNetV2 achieves 7.31 times compression while maintaining 81.46\% accuracy, making it highly suitable for mobile deployment scenarios where computational resources, memory bandwidth, and power consumption are critical constraints. ResNet-18 provides a middle ground with 2.19 times compression and higher absolute accuracy of 83.84\%, suitable for scenarios with moderate resource constraints but higher accuracy requirements. The availability of both students from our framework enables deployment flexibility, where practitioners can select the appropriate model based on their specific resource constraints and accuracy requirements.

\section{Discussion}

\subsection{Impact of Uncertainty Weighting}

The uncertainty-aware mechanism fundamentally changes how students learn from teacher predictions by introducing selectivity into the knowledge transfer process. When the teacher produces high-confidence predictions with low entropy, the corresponding uncertainty weight approaches unity, allowing students to fully leverage the teacher's knowledge for those samples. These confident predictions typically occur for prototypical examples that clearly exhibit class-defining characteristics, and they encode reliable patterns that students should prioritize learning.

Conversely, when the teacher encounters challenging or ambiguous samples that result in high-entropy predictions, the uncertainty weight diminishes, reducing the influence of potentially unreliable soft targets. Such uncertain predictions often arise for samples near decision boundaries, images with multiple overlapping objects, or unusual examples that deviate from typical class appearances. By reducing the emphasis on these uncertain predictions, students avoid learning potentially misleading patterns that could hurt generalization.

This selective learning strategy helps students focus their limited capacity on learning from confident teacher predictions, which are more likely to encode useful generalizable patterns rather than noise or artifacts. Simultaneously, the hard label loss component maintains a direct connection to ground truth supervision, ensuring that students do not become overly dependent on teacher predictions and retain the ability to correct teacher errors. This balance between selective teacher guidance and reliable ground truth supervision enables students to achieve superior performance compared to traditional knowledge distillation, which treats all teacher predictions uniformly regardless of confidence levels.

\subsection{Benefits of Peer Learning}

The heterogeneous peer architectures employed in our framework, ResNet-18 and MobileNetV2, bring complementary strengths that enhance collaborative learning. ResNet-18 with its deeper residual structure excels at capturing detailed spatial features and learning hierarchical representations through skip connections. Its relatively larger capacity allows it to model complex feature interactions and maintain fine-grained spatial information throughout the network depth. The residual connections enable learning of subtle patterns and relationships that might be difficult for smaller networks to capture.

MobileNetV2, designed for efficiency through depthwise separable convolutions, learns highly compact and efficient representations that emphasize the most discriminative features while minimizing computational overhead. The inverted residual structure and linear bottlenecks in MobileNetV2 force the network to learn compressed representations that capture essential information, discarding redundant or less important features. These architectural differences result in distinct learning trajectories and feature representations, where ResNet-18 might discover detailed texture patterns while MobileNetV2 focuses on high-level shape and structure.

Through peer distillation, ResNet-18 can benefit from MobileNetV2's focus on essential discriminative features, potentially improving its ability to generalize by learning to emphasize the most important patterns. Meanwhile, MobileNetV2 gains access to richer representations learned by ResNet-18's greater capacity, enabling it to capture some of the subtle patterns that might otherwise be lost in its compressed architecture. This mutual knowledge exchange enables both students to achieve higher accuracy than they would through isolated training with only teacher supervision, demonstrating the value of collaborative learning among heterogeneous architectures with complementary strengths.

\subsection{Limitations}

While our approach demonstrates significant improvements over baseline methods, several limitations warrant consideration. First, the dual-student framework requires training two networks simultaneously, resulting in increased training time compared to single-student baselines. Our experiments show approximately 1.63 times longer training duration, which may be prohibitive in scenarios with limited computational resources or tight development timelines. Organizations with constrained budgets or urgent deployment deadlines may find this additional training cost challenging to accommodate, despite the improved performance benefits.

Second, the framework introduces additional hyperparameters in the form of loss weights $\alpha$, $\beta$, and $\gamma$, which require careful tuning to achieve optimal performance. The balance between these components can significantly impact final accuracy, and finding the right configuration may require extensive experimentation across different datasets and architectures. Our experiments suggest that the optimal balance depends on factors such as dataset complexity, teacher quality, and student capacity, making it difficult to establish universal guidelines that work across all scenarios.

Third, our uncertainty estimation relies solely on prediction entropy, which captures distributional uncertainty but may not fully represent all aspects of prediction reliability. Entropy measures the spread of the probability distribution but does not distinguish between epistemic uncertainty arising from insufficient training data and aleatoric uncertainty inherent in the data itself. Alternative uncertainty measures such as Monte Carlo Dropout for epistemic uncertainty or evidential deep learning for quantifying different types of uncertainty could provide more comprehensive characterization of teacher confidence, potentially leading to further improvements in knowledge transfer quality.

Fourth, the framework has only been evaluated on image classification tasks with ImageNet-100, and its effectiveness on other domains or tasks remains to be demonstrated. Different tasks such as object detection, semantic segmentation, or natural language processing may exhibit different uncertainty patterns and knowledge transfer dynamics. Finally, our peer learning mechanism assumes that both students are trained simultaneously from the beginning, which may not be practical in scenarios where one student is already partially or fully trained. Investigating asynchronous training protocols or incorporating pre-trained students into the framework could broaden its applicability.

\subsection{Future Work}

Several promising directions emerge from this work that warrant further investigation. Extending the framework to incorporate more than two students could enable richer collaborative learning dynamics, where multiple heterogeneous architectures contribute diverse perspectives to the learning process. With three or more students of varying capacities and architectural designs, the peer learning mechanism could facilitate more complex knowledge exchange patterns, potentially uncovering synergies that are not accessible with only two students.

Exploring alternative uncertainty measures beyond entropy, such as Monte Carlo Dropout for epistemic uncertainty estimation or evidential deep learning for quantifying different types of uncertainty, could provide more nuanced guidance for knowledge transfer. These advanced uncertainty quantification methods might enable even more selective knowledge transfer by identifying not just uncertain predictions but also the source and nature of that uncertainty, allowing for more sophisticated weighting schemes.

Applying the uncertainty-aware dual-student framework to other domains such as natural language processing, speech recognition, or multimodal learning would demonstrate its generalizability beyond computer vision tasks. Each domain presents unique challenges and characteristics, and adapting our approach to these settings could reveal domain-specific insights and lead to further methodological innovations. For instance, in natural language processing, uncertainty might manifest differently due to the discrete and sequential nature of text, requiring adaptations to our entropy-based weighting scheme.

Investigating dynamic loss weight scheduling that adapts the balance between hard labels, teacher distillation, and peer learning throughout training could potentially improve convergence and final performance. Early in training, when students are still learning basic features, the optimal balance might differ from later stages when students are refining their decision boundaries. Adaptive weighting schemes could automatically adjust these balances based on training progress, validation performance, or uncertainty statistics.

Additionally, combining our approach with other model compression techniques such as pruning, quantization, or neural architecture search could lead to even more efficient models suitable for deployment on edge devices with severe resource constraints. Pruning could remove redundant parameters from students during or after distillation, while quantization could reduce precision requirements. Neural architecture search could automatically discover optimal student architectures that maximize accuracy under specific computational budgets when combined with our uncertainty-aware distillation framework.

Finally, investigating the transferability of learned uncertainty patterns across different datasets or tasks could provide insights into whether certain samples consistently produce uncertain predictions across different models and domains. Understanding these patterns could inform data collection strategies and help identify challenging samples that require additional annotation or augmentation.

\section{Conclusion}

This paper presented an uncertainty-aware dual-student knowledge distillation framework that leverages teacher prediction confidence to guide student learning effectively. By combining uncertainty-weighted teacher distillation with peer learning mechanisms, we achieved significant improvements over baseline knowledge distillation methods on ImageNet-100. ResNet-18 reached 83.84\% top-1 accuracy, representing a 2.04 percentage point improvement, and MobileNetV2 achieved 81.46\%, representing a 0.92 percentage point improvement over traditional single-student distillation.

Our approach demonstrates that selective knowledge transfer based on teacher confidence, combined with collaborative learning between heterogeneous architectures, can substantially improve model compression performance. The uncertainty-aware mechanism allows students to prioritize learning from reliable teacher predictions while maintaining robustness through ground truth supervision. The peer learning component enables mutual knowledge exchange between students with complementary architectural characteristics, further enhancing performance beyond what either student could achieve independently.

The proposed framework offers a practical solution for deploying efficient models on resource-constrained devices while maintaining competitive accuracy. With compression ratios of 2.19 times for ResNet-18 and 7.31 times for MobileNetV2 compared to the teacher, our approach enables flexible deployment strategies that can be tailored to specific resource constraints and accuracy requirements. The comprehensive ablation studies validate the contribution of each component, demonstrating that uncertainty weighting and peer learning each provide meaningful improvements beyond traditional knowledge distillation.

Future work will explore extensions to larger datasets such as full ImageNet, additional student architectures including vision transformers and more recent efficient architectures, and applications to other computer vision tasks beyond image classification. We anticipate that the principles underlying our framework, particularly uncertainty-aware knowledge transfer and heterogeneous peer learning, will prove valuable across a wide range of model compression scenarios and application domains.

\end{document}